\documentclass[10pt,twocolumn,letterpaper]{article}

\usepackage{cvpr}              
\definecolor{cvprblue}{rgb}{0.21,0.49,0.74}
\usepackage[pagebackref,breaklinks,colorlinks,allcolors=cvprblue]{hyperref}
\usepackage{multirow}
\usepackage{pifont}
\newcommand{\cmark}{\ding{51}}
\usepackage{caption}

\def\paperID{*****} 
\def\confName{CVPR}
\def\confYear{2026}

\title{APC: Transferable and Efficient Adversarial Point Counterattack\\ for Robust 3D Point Cloud Recognition}

\author{Geunyoung Jung$^1$ \qquad Soohong Kim$^1$ \qquad Inseok Kong$^2$ \qquad Jiyoung Jung$^1$\thanks{Corresponding author.} \\
$^1$Department of Artificial Intelligence, University of Seoul, South Korea \\
$^2$Department of Geo Informatics, University of Seoul, South Korea \\
{\tt\small \{gyjung975, shk00315, kong7963, jyjung\}@uos.ac.kr}
}

\begin{document}
\maketitle
\begin{abstract}
    The advent of deep neural networks has led to remarkable progress in 3D point cloud recognition, but they remain vulnerable to adversarial attacks. Although various defense methods have been studied, they suffer from a trade-off between robustness and transferability. We propose Adversarial Point Counterattack (APC) to achieve both simultaneously. APC is a lightweight input-level purification module that generates instance-specific counter-perturbations for each point, effectively neutralizing attacks. Leveraging clean–adversarial pairs, APC enforces geometric consistency in data space and semantic consistency in feature space. To improve generalizability across diverse attacks, we adopt a hybrid training strategy using adversarial point clouds from multiple attack types. Since APC operates purely on input point clouds, it directly transfers to unseen models and defends against attacks targeting them without retraining. At inference, a single APC forward pass provides purified point clouds with negligible time and parameter overhead. Extensive experiments on two 3D recognition benchmarks demonstrate that the APC achieves state-of-the-art defense performance. Furthermore, cross-model evaluations validate its superior transferability. The code is available at \small{\textcolor{violet}{\url{https://github.com/gyjung975/APC}}}.
\end{abstract}    
\section{Introduction}
\label{sec:intro}
    3D point cloud recognition is a fundamental task in computer vision, gaining increasing attention as 3D sensors like LiDAR provide reliable geometric information about complex real-world environments.
    Recent advances in deep neural networks (DNNs) have significantly improved the performance of point cloud recognition.
    However, DNNs remain highly vulnerable to adversarial attacks, in which subtle perturbations can deceive the networks into making incorrect predictions while undetectable to human observers.
    Since point cloud data have many safety-critical applications, ensuring the robustness of DNNs against adversarial attacks is crucial for the reliable and secure deployment of 3D vision applications.

    Existing adversarial attacks on point clouds generally fall into three strategies: adding, dropping, and shifting points.
    Early approaches~\cite{3d_adv, minimal} generate adversarial examples by inserting adversarial points or clusters into the original input.
    The dropping strategy~\cite{drop} constructs a point saliency map to determine the importance of each point and removes the points identified as important.
    Point shifting is another widely used strategy, where the coordinates of points are perturbed to induce misclassification.
    For instance, C\&W-based~\cite{3d_adv} and PGD-based~\cite{icip} attacks iteratively refine perturbations to produce stronger adversarial examples.
    However, many of these attacks suffer from limited imperceptibility, making them easier to detect.
    To alleviate this issue, KNN~\cite{knn} incorporates distance constraints between adversarial and neighboring points, thereby preventing removal by defense methods.
    More recently, shape- and geometry-aware attacks~\cite{sia, geo, hit} are introduced to further enhance imperceptibility by preserving local structure and curvature. 
    Notably, HiT~\cite{hit} hides perturbations within geometrically complex areas, guided by saliency and imperceptibility scores.

    Since the robustness of DNNs against adversarial attacks is critical in real-world applications, a variety of defense methods are proposed.
    Existing approaches can be categorized into two types: input-level~\cite{dup, gvg, if, pointguard, causalpc} and model-level~\cite{icip, at, pointdrop, bench} defenses.
    Input-level defenses directly manipulate adversarial examples in data space. 
    The most straightforward methods are Simple Random Sampling (SRS) and Statistical Outlier Removal (SOR)~\cite{dup}, which remove random points and statistical outlier points, respectively.
    In a more sophisticated approach, IF-Defense~\cite{if} optimizes point coordinates at inference time to restore the underlying surface.
    Inspired by randomized smoothing~\cite{randomized, rse} in the 2D domain, CausalPC~\cite{causalpc} ensembles outputs from multiple augmented adversarial examples.
    Model-level defenses aim to improve the robustness of the model itself, rather than manipulating adversarial inputs. 
    Adversarial training is the most widely adopted paradigm~\cite{fgsm, icip, at, pointdrop}, where adversarial examples are incorporated into training to enhance inherent robustness.
    Recently, hybrid training~\cite{bench} leverages adversarial examples from multiple attacks, achieving strong defense performance.

    In general, the two types of defenses exhibit complementary strengths and weaknesses.
    By operating purely in the data space, input-level defenses are model-agnostic and offer high transferability across models. 
    However, their robustness remains relatively weak, as it is only indirectly induced by input manipulation rather than explicit defensive objectives.
    In contrast, model-level defenses achieve strong robustness but lack transferability and are less practical, as each model must be retrained from scratch with a specific robust objective.
    
    To address the drawbacks of both types of defense simultaneously, we propose Adversarial Point Counterattack (APC), a lightweight input-level purification module.
    APC takes adversarial examples as input and generates instance-specific counter-perturbations that neutralize the effects of attacks at each point.
    By adding these counter-perturbations point-wise to adversarial examples, we obtain purified examples.
    APC is trained on clean-adversarial pairs with the objective of restoring adversarial examples to their clean counterparts. 
    To further improve restoration capability, two consistency losses are employed: one enforcing coordinate consistency in the data space and the other ensuring feature consistency in the feature space.
    Additionally, following ~\cite{bench}, we adopt a hybrid training strategy to enable universal defense.
    Once trained with adversarial examples from multiple attack types, APC achieves robustness against both seen and unseen adversarial attacks.
    While APC outperforms all existing model-level defenses, it also exhibits strong transferability due to its input-level design. 
    Moreover, APC is lightweight and efficient, introducing negligible additional cost at inference time.

    Extensive experiments on 3D recognition across diverse datasets and models demonstrate the effectiveness of APC.
    APC achieves state-of-the-art defense performance, surpassing both input- and model-level defenses.
    To evaluate its transferability, we conduct cross-model generalization experiments, where APC significantly outperforms existing input-level defenses, highlighting its strong transferability.

    The main contributions can be summarized as follows:
    \begin{itemize}
        \item We propose APC, a lightweight input-level defense module that generates counter-perturbations to counteract adversarial attacks.
        Trained on clean–adversarial pairs with two consistency losses, APC effectively purifies adversarial examples.

        \item Through hybrid training and its input-level design, a single APC generalizes to unseen attacks and models, providing consistent robustness in both cross-attack and cross-model settings.

        \item APC achieves state-of-the-art defense performance on 3D recognition, outperforming both input-level and model-level defenses. Cross-model evaluations further demonstrate its strong transferability.
    \end{itemize}
\section{Related Works}
\label{sec:related}
    \subsection{Deep Learning on 3D Point Cloud Recognition}
        3D point clouds are widely used in various real-world applications such as autonomous driving and robotics, but their irregular and unordered nature poses significant challenges for effective processing.
        PointNet~\cite{pointnet} pioneers the direct processing of raw point clouds.
        PointNet++~\cite{pointnet++} further captures local geometric structures at multi-scale through a hierarchical architecture.
        Recent works introduce 3D convolutions~\cite{pointcnn, pointconv, pointcnn2, kpconv, relation_cnn} and graph-based models~\cite{dynamic_edge_filters, minimg_local, grid_gcn, dgcnn}.
        DGCNN~\cite{dgcnn} is a representative graph-based model that leverages dynamic graph convolutions, constructing local graphs adaptively in both coordinate and feature spaces. 
        In this paper, we focus on PointNet, PointNet++, and DGCNN, as they are widely adopted across diverse 3D tasks.

    \begin{figure*}[!ht]
    \centering
        \includegraphics[width=1.\linewidth]{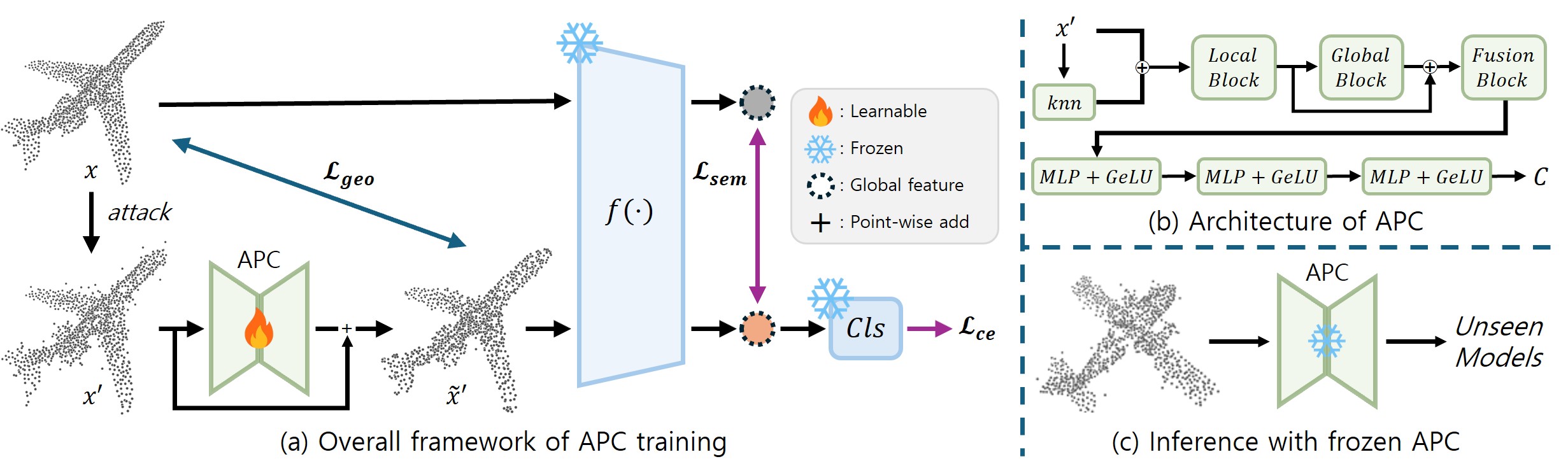}
    \caption{Overall pipeline of APC training and inference. Given an adversarial point cloud $x'$ created by attacking a clean point cloud $x$, APC generates counter-perturbations $C$. By point-wise adding $C$ to $x'$, we obtain the purified point cloud $\tilde{x}'$. $f(\cdot)$ and $Cls$ denote feature extractor and classifier of the victim model, respectively. Once trained, APC can be transferred to any model, as depicted in (c).}
    \label{fig:framework}
    \end{figure*}

    \subsection{3D Adversarial Attacks and Defenses}
        \paragraph{Attacks.}
            Following the works in the 2D domain~\cite{2dadv, cw}, adversarial attacks are extended to 3D point clouds through three main strategies: adding, dropping, and shifting points. 
            Add-based attacks~\cite{3d_adv,minimal} generate adversarial examples by inserting new points or clusters into the original point clouds. 
            Conversely, Saliency~\cite{drop} removes points with high saliency, using a saliency map that assigns importance scores to each point. 
            The most widely explored approach is shifting, where the coordinates of existing points are minimally perturbed. 
            C\&W-based~\cite{3d_adv} and PGD-based~\cite{icip} attacks iteratively optimize per-point perturbations. 
            Since imperceptibility is a crucial aspect for 3D attacks, recent works explicitly consider it.
            To reduce outlier-like perturbations, KNN~\cite{knn} incorporates a KNN distance constraint to encourage local compactness. 
            More advanced shape- and geometry-aware approaches~\cite{shape, sia, geo, hit} improve imperceptibility by preserving the surface structure and local curvature.
            Additionally, HiT~\cite{hit} conceals perturbations within less perceptually sensitive regions, based on saliency and imperceptibility scores.
            AdvPC~\cite{advpc} employs an auto-encoder to generate more transferable adversarial examples, thereby improving the attack transferability.

        \paragraph{Defenses.}
            Existing defense methods can be categorized into two main streams: input-level and model-level.
            Input-level methods operate directly in the data space by transforming input point clouds. 
            The simplest methods include Simple Random Sampling (SRS) and Statistical Outlier Removal (SOR)~\cite{dup}, which remove random or outlier points.
            DUP-Net~\cite{dup} upsamples adversarial examples to restore missing details and refine surfaces using the reconstruction model PU-Net~\cite{pu_net}.
            For more effective defense, IF-Defense~\cite{if} performs instance-specific iterative defense by optimizing point coordinates during inference. 
            It reconstructs the surface via an implicit function network, such as ConvONet~\cite{convonet}. 
            Another line of defense utilizes randomized smoothing~\cite{pointguard, causalpc}, which averages outputs from multiple augmented adversarial examples. However, they require a large number of forward passes (100--10,000) and additional training of victim models.

            Model-level methods enhance robustness by retraining each victim model from scratch. 
            A common strategy is adversarial training~\cite{icip, at}, which augments training data with adversarial examples. 
            Recently, recognizing that training with a single attack type is insufficient, hybrid training~\cite{bench} incorporates multiple attack types.

    \subsection{3D Point Cloud Denoising}
        3D point cloud denoising aims to restore clean geometry from noisy inputs, typically achieved via coordinate-level reconstruction using local geometric priors or learned representations. 
        IterativePFN~\cite{iterativepfn} performs iterative filtering through stacked denoising modules with dynamic graph convolution. 
        PD-LTS~\cite{pdlts} leverages latent space to uncover noise components and extract clean latent code for denoising. 
        U-CAN~\cite{u_can} adopts an unsupervised multi-step denoising strategy based on point-wise noise-to-noise matching.

        Although input-level adversarial defenses resemble denoising in modifying point coordinates, denoising focuses on geometric fidelity, whereas adversarial defense requires preserving task-relevant semantics. 

\section{Method}
\label{sec:method}
    We first introduce the architecture and purification process of our Adversarial Point Counterattack (APC), followed by the training objectives in \cref{sec:apc}. 
    Then, \cref{sec:hybrid} presents a hybrid training strategy that enables universal defense by jointly leveraging multiple attack types.
    Finally, \Cref{sec:inference} discusses the inference process of APC, highlighting its transferability and efficiency.
    The overall framework of APC is illustrated in \cref{fig:framework}.

    \subsection{Adversarial Point Counterattack}
    \label{sec:apc}
        APC is an input-level defense method that takes an adversarial point cloud as input and generates per-point counter-perturbations to neutralize attacks.
        The purified point cloud is obtained by adding these counter-perturbations point-wise.
        Leveraging a dataset of clean–adversarial pairs, APC is trained to transform adversarial point clouds into their clean counterparts.

        \subsubsection{Distribution-aware Counter-Perturbation}
            APC uses an encoder-decoder architecture, where the encoder extracts point features and the decoder generates per-point counter-perturbations. 
            The encoder consists of three blocks: local, global, and fusion. 
            The local and global blocks employ an MLP followed by max pooling to capture fine-grained local geometry and broader global context, respectively. 
            The fusion block, implemented as an MLP, integrates these features to enhance the robustness of point features against adversarial attacks. 

            To effectively counteract attacks at individual point level, we begin with $k$-nearest neighbor (KNN) aggregation, which leverages local geometric information to suppress localized noise.
            Given an adversarial point cloud $x' \in \mathbb{R}^{N\times 3}$ with $N$ points, we select $k$-nearest points for each point.
            Each point and its neighbors are concatenated and passed through local block $g^{local}$ to generate local point features:
            \begin{equation}
                P = KNN(x'),\quad P \in \mathbb{R}^{N \times k \times 3} \\ 
            \end{equation}
            \begin{equation}
                L = g^{local}([\operatorname{repeat}_k(x');P]),\quad L \in \mathbb{R}^{N \times d},
            \end{equation}
            where $\operatorname{repeat}_k(x')$ denotes repeating each point $k$ times along the second dimension, and $d$ is the feature dimension.
            In addition, we further utilize a global feature to encode overall shape information via the global block $g^{global}$:
            \begin{equation}
                G = g^{global}(L),\quad G \in \mathbb{R}^{1 \times d}
            \end{equation}
            This global feature provides a holistic representation of the object and remains relatively stable against local perturbations.
            Finally, we fuse the local point features and the global features through $g^{fusion}$ to obtain final point features:
            \begin{equation}
                E = g^{fusion}([L;\operatorname{repeat}_N(G)]),\quad E \in \mathbb{R}^{N \times d},
            \end{equation}
            where $G$ is repeated $N$ times along the first dimension to match the shape of $L$.
            This fusion contextualizes each point with both its local neighborhood and the overall shape, thereby improving robustness against adversarial attacks.

            The fused point features are then decoded into counter-perturbations $C$ by a decoder $h$, composed of a 3-layer MLP with GeLU activations.
            By adding these counter-perturbations to the adversarial point cloud in a point-wise manner, we obtain the purified point cloud $\tilde{x}'$:
            \begin{equation}
                C = h(E),\quad C \in \mathbb{R}^{N \times 3}
            \end{equation}
            \begin{equation}
                \tilde{x}' = x' + C
            \end{equation}
            The resulting purified point cloud is subsequently fed into the models for downstream tasks.

        \subsubsection{Optimization}
            The optimal goal of input-level defense is to recover the original clean examples from their adversarial counterparts. 
            Therefore, we construct a dataset of clean-adversarial pairs, which provides direct supervision for restoring clean examples. 
            Using this paired dataset, APC is trained with losses that enforce both geometric and semantic consistency.
            
            \paragraph{Data-Space Geometric Consistency Loss.}
                To ensure that a purified example retains geometric structure of its clean counterpart, we introduce a geometric consistency loss. 
                The purified example $\tilde{x}'$ is constrained to align with the clean example $x$ at the coordinate level. 
                We achieve this by minimizing the Chamfer distance~\cite{cham}, which measures how far two point clouds are from each other, between $\tilde{x}'$ and $x$. 
                It is calculated by averaging the distances of all nearest point pairs.
                The geometric consistency loss is defined as:
                \begin{equation}
                    \mathcal{L}_{geo} = \frac{1}{\Vert \tilde{x}' \Vert_0}\sum_{p'\in \tilde{x}'}\min_{p \in x}\Vert p'-p \Vert^2_2
                \end{equation}
    
            \paragraph{Feature-Space Semantic Consistency Loss.}
                In addition to geometric consistency, restoring high-level semantics is also important for recognition. 
                To this end, we further introduce a semantic consistency loss between the features of the purified and clean examples to ensure that the purified features retain essential semantics of the clean ones. 
                Specifically, we use mean squared error (MSE) to encourage the global features of $\tilde{x}'$ and $x$ to be similar, defined as:
                \begin{equation}
                    \mathcal{L}_{sem} = \frac{1}{d}\sum^d_{i=1}(f(\tilde{x}')_i - f(x)_i)^2,
                \end{equation}
                where $f(\cdot)$ denotes the feature extractor of the model, and $f(\tilde{x}')$ and $f(x)$ represent the global features of the purified example and its corresponding clean example, respectively.

                By jointly optimizing losses in the data space and feature space, the purification process of APC not only rectifies local coordinate perturbations but also restores high-level semantic information.

            \noindent\textbf{Final Loss.}
                The final loss combines the two consistency terms with a standard cross-entropy loss as follows:
                \begin{equation}
                    \mathcal{L} = \mathcal{L}_{ce} + \alpha\cdot\mathcal{L}_{geo} + \beta\cdot\mathcal{L}_{sem}, 
                \end{equation}
                where $\alpha$ and $\beta$ are hyperparameters that balance the contributions of each consistency loss.

    \begin{figure}[!t]
    \centering
        \includegraphics[width=1.\linewidth]{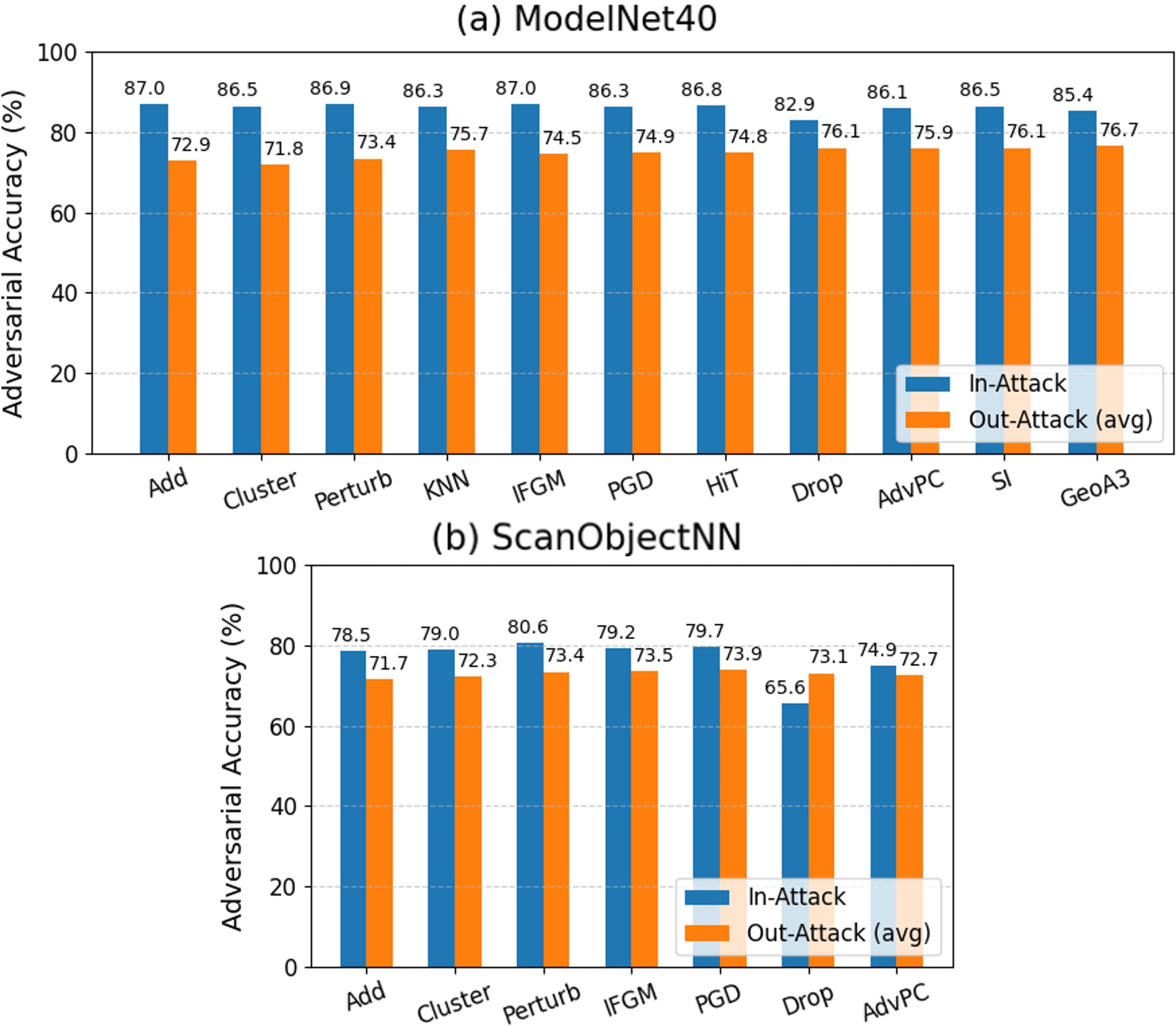}
    \caption{Adversarial accuracies of IAPC on in-attack and out-attack (avg) across two datasets. In-attacks are almost completely defended, but the performance on out-attacks is relatively low.}
    \label{fig:iapc}
    \end{figure}

    \begin{table*}[!t]
    \centering
    \caption{Classification results (\%) of adversarial examples of various attacks on ModelNet40 with three victim models. The experiments are conducted under a white-box setting and with untargeted attacks. \textbf{Bold} indicates the best performance. Overall, APC consistently achieves the highest robustness across different models and attacks.}
    \label{tab:cls_mn}
    \resizebox{\textwidth}{!}{%
    \begin{tabular}{lc|cccccccccccc}
        \toprule
        Model & Defense & Add & Cluster & Perturb & KNN & IFGM & PGD & HiT & Drop & AdvPC & SI & GeoA3 & Avg. \\
        \midrule
        \multirow{9}{*}{PointNet} & No Defense & 0.0 & 0.2 & 0.0 & 4.7 & 12.5 & 11.7 & 0.9 & 36.1 & 0.0 & 0.0 & 0.0 & 6.0 \\
         & SRS        & 81.5 & 74.7 & 79.7 & 29.7 & 80.5 & 79.7 & 17.7 & 34.6 & 24.4 & 54.3 & 55.5 & 55.7 \\
         & SOR        & 80.9 & 84.3 & 85.4 & 65.3 & 84.6 & 84.4 & 16.7 & 38.6 & 44.7 & 72.7 & 65.8 & 65.8 \\
         & DUP-Net    & 82.4 & 83.3 & 84.8 & 75.3 & 84.7 & 84.5 & 31.1 & 44.6 & 69.3 & 79.6 & 76.2 & 72.3 \\
         & IF-Defense & 84.2 & 84.4 & 84.2 & 80.4 & 85.0 & 83.8 & 83.6 & 58.3 & 79.2 & 82.0 & 81.3 & 80.6 \\
         & CausalPC   & 84.4 & 84.3 & 84.2 & 80.6 & 84.0 & 83.9 & 38.3 & 44.9 & 76.6 & 82.2 & 78.9 & 74.8 \\
         & AT         & 85.1 & 80.7 & 86.0 & 68.4 & 86.0 & 85.8 & 51.6 & 59.1 & 76.2 & 77.1 & 81.5 & 76.1 \\
         & HT         & 84.3 & 84.2 & 84.7 & 79.6 & 84.8 & 83.7 & 62.5 & 75.5 & 78.9 & 81.8 & 81.1 & 80.1 \\
         & APC        & \textbf{86.4} & \textbf{86.3} & \textbf{86.6} & \textbf{84.0} & \textbf{86.7} & \textbf{86.7} & \textbf{85.9} & \textbf{79.4} & \textbf{81.5} & \textbf{85.0} & \textbf{83.3} & \textbf{84.7} \\
        \midrule
        \midrule
        \multirow{9}{*}{PointNet++} & No Defense & 84.4 & 15.6 & 80.9 & 1.9 & 51.1 & 48.2 & 65.3 & 58.8 & 13.2 & 1.8 & 33.7 & 41.4 \\
         & SRS        & 86.9 & 70.3 & 85.5 & 24.7 & 78.9 & 78.8 & 62.2 & 54.4 & 44.8 & 56.9 & 74.1 & 65.2 \\
         & SOR        & \textbf{88.0} & 87.5 & 87.5 & 50.4 & 84.6 & 84.8 & 62.5 & 60.5 & 59.1 & 80.2 & 80.6 & 75.1 \\
         & DUP-Net    & \textbf{88.0} & 86.9 & \textbf{87.6} & 59.8 & 85.7 & 85.6 & 65.7 & 64.0 & 64.1 & 82.2 & 82.1 & 77.4 \\
         & IF-Defense & 86.9 & 85.9 & 87.4 & 73.7 & 86.2 & 85.9 & 86.7 & 70.7 & 76.2 & 83.5 & 82.8 & 82.3 \\
         & AT         & 87.0 & 83.9 & 87.4 & 52.1 & 86.8 & 86.8 & 74.0 & 72.0 & 71.4 & 76.4 & 85.4 & 78.5 \\
         & HT         & 87.1 & 85.0 & 87.2 & 79.6 & \textbf{87.8} & 86.2 & 65.4 & 84.1 & 82.7 & 84.1 & \textbf{87.7} & 83.3 \\
         & APC        & 87.6 & \textbf{87.5} & 86.8 & \textbf{83.3} & 87.0 & \textbf{86.7} & \textbf{86.5} & \textbf{75.9} & \textbf{82.6} & \textbf{85.9} & 84.6 & \textbf{84.9} \\
        \midrule
        \midrule
        \multirow{9}{*}{DGCNN} & No Defense & 6.7 & 5.7 & 4.3 & 1.9 & 32.3 & 30.6 & 19.7 & 58.1 & 0.9 & 0.9 & 0.0 & 14.6 \\
         & SRS        & 80.9 & 66.7 & 69.1 & 29.9 & 71.8 & 71.2 & 49.7 & 24.4 & 28.7 & 48.8 & 56.5 & 54.3 \\
         & SOR        & 87.2 & 86.0 & 87.8 & 31.7 & 87.2 & 87.1 & 47.2 & 62.6 & 41.2 & 74.9 & 72.9 & 69.6 \\
         & DUP-Net    & 52.3 & 52.9 & 51.8 & 34.7 & 51.8 & 51.2 & 19.5 & 37.0 & 30.3 & 42.2 & 40.2 & 42.2 \\
         & IF-Defense & 87.1 & 85.5 & 87.0 & 70.3 & 86.9 & 85.6 & 84.1 & 68.8 & 75.0 & 82.4 & 82.7 & 81.4 \\
         & CausalPC   & 84.4 & 83.7 & 83.3 & 76.9 & 84.0 & 83.7 & 47.7 & 61.1 & \textbf{78.2} & 81.2 & 82.7 & 77.0 \\
         & AT         & \textbf{87.3} & 84.0 & 87.0 & 51.9 & 87.3 & 87.1 & 72.8 & 67.4 & 62.3 & 81.9 & 84.8 & 77.6 \\
         & HT         & 78.4 & 78.9 & \textbf{87.9} & 52.1 & \textbf{87.9} & 87.1 & 72.9 & 68.9 & 60.9 & \textbf{83.0} & \textbf{85.4} & 76.7 \\
         & APC        & 86.9 & \textbf{86.8} & 86.7 & \textbf{77.3} & 86.8 & \textbf{87.1} & \textbf{87.4} & \textbf{75.4} & 76.7 & 82.4 & 83.9 & \textbf{83.4} \\
        \bottomrule
    \end{tabular}%
    }
    \end{table*}

    \subsection{Universal Counterattack}
    \label{sec:hybrid}
        Training APC with only a single type of attack may limit its generalizability.
        \Cref{fig:iapc} illustrates the cross-attack performance of APC trained on a specific attack, denoted as Independent-APC (IAPC).
        Here, in-attack and out-attack (avg) refer to the performance on the attack used for training and the average performance over the remaining attacks, respectively.
        Each IAPC exhibits strong robustness in the in-attack setting, achieving an average accuracy of 86.2\% on ModelNet40~\cite{modelnet}, which is close to the clean accuracy of the vanilla model (87.6\%). 
        However, its performance degrades significantly when evaluated on unseen attacks.
        This suggests that training with a single attack leads to overfitting and hinders generalization across diverse attack patterns.
        Note that although IAPC's out-attack performance lags behind its in-attack result, it still shows solid robustness on ModelNet40 (74.8\%) and even achieves a state-of-the-art performance on ScanObjectNN (73.0\%), as shown in \cref{tab:cls_mn,tab:cls_sonn}.

        Following~\cite{bench}, we adopt a hybrid training strategy to alleviate this limitation. 
        We train APC jointly on adversarial examples generated from multiple attack types. 
        Specifically, we select one attack from each of the following categories: shifting, shape-aware shifting, and dropping. 
        By exposing APC to heterogeneous attack patterns during training, it learns a more universal purification ability. 
        As we demonstrate in \cref{sec:exp}, this approach substantially improves out-attack robustness while maintaining strong in-attack performance.

    \subsection{Inference with Frozen APC}
    \label{sec:inference}
        In this section, we describe the inference process using the trained APC, highlighting its transferability and efficiency.
        As shown in part (c) of \cref{fig:framework}, APC can be transferred in a plug-and-play manner to various models, since it performs input-level purification.
        With a frozen APC, adversarial examples targeting models other than the one used for APC training can also be defended.
        For instance, adversarial examples crafted to deceive DGCNN~\cite{dgcnn} can be purified by an APC trained with PointNet~\cite{pointnet}.

        We emphasize that, through its superior transferability and universal defense capability enabled by hybrid training, APC can be applied to a wide range of unseen models and remains effective against diverse attacks. 
        Moreover, it is lightweight, containing only 0.2M parameters (7\% of PointNet), and extremely efficient, as purification requires only a single APC forward pass.

\section{Experiments}
\label{sec:exp}
    \begin{table}[!t]
    \scriptsize
    \centering
    \caption{Classification results (\%) of adversarial examples on ScanObjectNN OBJ\_ONLY split.}
    \label{tab:cls_sonn}
    \resizebox{0.475\textwidth}{!}{%
    \begin{tabular}{l@{\hskip 5pt}c|c@{\hskip 5pt}c@{\hskip 5pt}c@{\hskip 5pt}c@{\hskip 5pt}c@{\hskip 5pt}c@{\hskip 5pt}c@{\hskip 5pt}c}
        \toprule
        Model & Defense & Add & Cluster & Perturb & IFGM & PGD & Drop & AdvPC & Avg. \\
        \midrule
        \multirow{9}{*}{PointNet} & No Defense & 0.3 & 0.3 & 0.0 & 10.0 & 9.6 & 45.9 & 0.0 & 9.4 \\
         & SRS        & 70.9 & 69.8 & 70.9 & 73.8 & 73.4 & 46.3 & 22.9 & 61.2 \\
         & SOR        & 77.7 & 79.0 & 78.6 & 78.7 & 78.6 & 47.5 & 46.1 & 69.4 \\
         & DUP-Net    & 77.8 & 78.6 & 78.7 & 79.1 & 78.3 & 51.1 & 57.6 & 71.6 \\
         & IF-Defense & 44.2 & 43.0 & 43.0 & 43.2 & 43.2 & 38.7 & 39.9 & 42.2 \\
         & CausalPC   & 75.6 & 74.2 & 73.5 & 72.8 & 73.8 & 50.1 & 61.8 & 68.8 \\
         & AT         & 77.1 & 75.2 & 77.6 & 77.2 & 78.1 & 63.0 & 62.0 & 72.9 \\
         & HT         & 72.2 & 66.4 & 75.9 & 75.9 & 74.8 & \textbf{64.7} & 61.9 & 70.3 \\
         & APC        & \textbf{80.0} & \textbf{80.0} & \textbf{80.6} & \textbf{80.4} & \textbf{80.4} & 59.9 & \textbf{71.3} & \textbf{76.1} \\
        \midrule
        \midrule
        \multirow{9}{*}{PointNet++} & No Defense & 77.7 & 17.3 & 73.1 & 31.8 & 28.4 & 63.0 & 14.1 & 43.7 \\
         & SRS        & 82.7 & 66.9 & 83.4 & 79.5 & 78.1 & 64.3 & 54.7 & 72.8 \\
         & SOR        & 85.5 & 85.2 & 85.7 & 84.1 & 83.1 & 65.4 & 64.7 & 79.1 \\
         & DUP-Net    & 85.8 & 85.3 & \textbf{87.4} & 82.9 & 84.1 & 68.5 & 67.1 & 80.2 \\
         & IF-Defense & 48.0 & 47.8 & 47.3 & 46.8 & 47.2 & 44.4 & 40.4 & 46.0 \\
         & AT         & 84.5 & 76.4 & 83.6 & 81.5 & 82.1 & 71.9 & 69.5 & 78.5 \\
         & HT         & 84.0 & 65.5 & 82.9 & 80.2 & 79.7 & 73.8 & 64.8 & 75.8 \\
         & APC        & \textbf{85.7} & \textbf{85.3} & 85.2 & \textbf{84.9} & \textbf{85.2} & \textbf{74.9} & \textbf{79.0} & \textbf{82.9} \\
        \midrule
        \midrule
        \multirow{9}{*}{DGCNN} & No Defense & 4.3 & 5.5 & 3.4 & 21.3 & 23.6 & 62.8 & 0.5 & 17.4 \\
         & SRS        & 77.1 & 67.8 & 72.9 & 76.6 & 75.2 & 44.7 & 45.1 & 65.6 \\
         & SOR        & 82.7 & \textbf{86.5} & 85.3 & 84.5 & 86.0 & 65.4 & 51.8 & 77.5 \\
         & DUP-Net    & 76.0 & 73.8 & 75.2 & 77.4 & 75.7 & 64.9 & 58.0 & 71.6 \\
         & IF-Defense & 52.6 & 53.2 & 54.2 & 54.0 & 52.1 & 46.3 & 41.8 & 50.6 \\
         & CausalPC   & 80.7 & 81.1 & 82.8 & 81.9 & 81.1 & 68.3 & \textbf{76.6} & 78.9 \\
         & AT         & 84.5 & 81.9 & 83.5 & 85.7 & 85.9 & 71.1 & 62.3 & 79.3 \\
         & HT         & 80.7 & 76.0 & 81.4 & 84.8 & 85.3 & \textbf{79.3} & 68.5 & 79.4 \\
         & APC        & \textbf{86.2} & 85.5 & \textbf{86.7} & \textbf{87.4} & \textbf{87.1} & 76.2 & 63.9 & \textbf{81.9} \\
        \bottomrule
    \end{tabular}%
    }
    \end{table}

    \begin{table*}[!t]
    \centering
    \caption{Adversarial accuracies (\%) on ModelNet40 in the cross-model transfer setting. APC is trained with the source model and evaluated with target models without retraining. The results are averages of performance across all attacks. APC exhibits strong transferability, consistently outperforming baseline defenses across diverse source–target pairs.}
    \label{tab:cross_mn}
    \resizebox{\textwidth}{!}{%
    \begin{tabular}{l|ccc|ccc|ccc}
        \toprule
        \multirow{2}{*}{Defense} & 
        Source & \multicolumn{2}{c|}{Target} & 
        Source & \multicolumn{2}{c|}{Target} & 
        Source & \multicolumn{2}{c}{Target} \\
        \cmidrule(lr){2-2} \cmidrule(lr){3-4} \cmidrule(lr){5-5} \cmidrule(lr){6-7} \cmidrule(lr){8-8} \cmidrule(lr){9-10}
         & PointNet & PointNet++ & DGCNN & PointNet++ & PointNet & DGCNN & DGCNN & PointNet & PointNet++ \\
        \midrule
        SRS        & 55.7 & 65.2 & 54.3 & 65.2 & 55.7 & 54.3 & 54.3 & 55.7 & 65.2 \\
        SOR        & 65.8 & 75.1 & 69.6 & 75.1 & 65.8 & 69.6 & 69.6 & 65.8 & 75.1 \\
        DUP-Net    & 72.3 & 77.4 & 42.2 & 77.4 & 72.3 & 42.2 & 42.2 & 72.3 & 77.4 \\
        IF-Defense & 80.6 & 82.3 & 81.4 & 82.3 & \textbf{80.6} & 81.4 & 81.4 & \textbf{80.6} & 82.3 \\ 
        APC        & \textbf{84.7} & \textbf{83.1} & \textbf{81.5} & \textbf{84.9} & 77.2 & \textbf{81.6} & \textbf{83.4} & 77.2 & \textbf{83.0} \\
        \bottomrule
    \end{tabular}%
    }
    \end{table*}

    \begin{table*}[!t]
    \centering
    \caption{Adversarial accuracies (\%) on ScanObjectNN OBJ\_ONLY split in the cross-model transfer setting.}
    \label{tab:cross_sonn}
    \resizebox{\textwidth}{!}{%
    \begin{tabular}{l|ccc|ccc|ccc}
        \toprule
        \multirow{2}{*}{Defense} & 
        Source & \multicolumn{2}{c|}{Target} & 
        Source & \multicolumn{2}{c|}{Target} & 
        Source & \multicolumn{2}{c}{Target} \\
        \cmidrule(lr){2-2} \cmidrule(lr){3-4} \cmidrule(lr){5-5} \cmidrule(lr){6-7} \cmidrule(lr){8-8} \cmidrule(lr){9-10}
         & PointNet & PointNet++ & DGCNN & PointNet++ & PointNet & DGCNN & DGCNN & PointNet & PointNet++ \\
        \midrule
        SRS        & 61.2 & 72.8 & 65.6 & 72.8 & 61.2 & 65.6 & 65.6 & 61.2 & 72.8 \\
        SOR        & 69.4 & 79.1 & 77.5 & 79.1 & 69.4 & 77.5 & 77.5 & 69.4 & 79.1 \\
        DUP-Net    & 71.6 & 80.2 & 71.6 & 80.2 & 71.6 & 71.6 & 71.6 & 71.6 & 80.2 \\
        IF-Defense & 42.2 & 46.0 & 50.6 & 46.0 & 42.2 & 50.6 & 50.6 & 42.2 & 46.0 \\ 
        APC        & \textbf{76.1} & \textbf{81.6} & \textbf{80.2} & \textbf{82.9} & \textbf{72.1} & \textbf{79.0} & \textbf{81.9} & \textbf{74.1} & \textbf{81.3} \\
        \bottomrule
    \end{tabular}%
    }
    \end{table*}

    \subsection{Experimental Settings}
        \paragraph{Datasets and Models.}
            We conduct experiments on ModelNet40 (MN40)~\cite{modelnet} and ScanObjectNN (SONN)~\cite{scanobjectnn} datasets. 
            MN40 is a synthetic dataset consisting of complete and noise-free point clouds. It contains 12,311 samples across 40 categories, with 9,843 and 2,468 samples for the train and test sets, respectively.
            SONN is a real-scanned dataset with 15 categories, characterized by substantial missing points and noise. The dataset is divided into three splits according to difficulty level (OBJ\_ONLY, OBJ\_BG, and PB), and we report results on the OBJ\_ONLY split.
            Additional results on the OBJ\_BG and PB splits are provided in the supplementary materials.
            For both datasets, we uniformly sample 1,024 points from each point cloud.
            For victim models, we adopt three commonly used point cloud classification models: PointNet~\cite{pointnet}, PointNet++~\cite{pointnet++}, and DGCNN~\cite{dgcnn}.

        \paragraph{Baseline Attacks.}
            We use eleven representative adversarial attacks: two adding attacks (Add~\cite{3d_adv}, Cluster~\cite{3d_adv}), one dropping attack (Drop~\cite{drop}), and seven shifting attacks (Perturb~\cite{3d_adv}, KNN~\cite{knn}, IFGM, PGD~\cite{icip}, HiT~\cite{hit}, AdvPC~\cite{advpc}, SI~\cite{sia}, and GeoA3~\cite{geo}).
            All attacks are implemented with untargeted, white-box setting.
            From the attacker's perspective, untargeted and white-box attacks are generally stronger than targeted and black-box attacks.
            Therefore, the untargeted under white-box setting represents the upper bound of attack strength and the most challenging scenario to defend.

        \paragraph{Baseline Defenses.}
            For defense baselines, we consider seven methods: five input-level defenses (SRS, SOR~\cite{dup}, DUP-Net~\cite{dup}, IF-Defense~\cite{if}, and CausalPC~\cite{causalpc}), and two model-level defenses (adversarial training (AT)~\cite{at} and hybrid training (HT)~\cite{bench}). 
            Since CausalPC has no official implementation, we re-implement it with 1,024 points following the paper.
            AT is trained with a mixture of PGD adversarial examples and clean examples.
            Unlike AT, HT leverages multiple attack types---KNN, HiT, and Drop---representing shifting, shape-aware shifting, and dropping attacks, respectively.

        \paragraph{Implementation Details.}
            For APC training, we adopt the combination of KNN, HiT, and Drop attacks together with clean examples, same as HT.
            To improve data efficiency, we only use 30\% of the training examples for each attack, as well as clean examples.
            The victim models are kept frozen while only the APC is updated, and clean-adversarial pairs are constructed from the training split of each dataset.
            Following existing input-level defenses, we also pre-process input data with SOR before feeding it into APC.

    \subsection{3D Point Cloud Recognition}
        \Cref{tab:cls_mn} summarizes the adversarial classification accuracies on MN40 under eleven different attacks across three victim models. 
        APC consistently performs either best or second-best across nearly all attacks and models. 
        On PointNet, it outperforms all baselines, including both input-level and model-level defenses, achieving a state-of-the art (SOTA) average accuracy of 84.7\%.
        Also on PointNet++ and DGCNN, it achieves average accuracies of 84.9\% and 83.4\%, respectively, again surpassing existing defenses by a clear margin.

        Unlike MN40, SONN does not provide the normal vector of each point. Therefore, we only consider attacks that can be implemented with coordinates alone.
        For HT and APC training, the combination of attacks becomes Perturb, PGD, and Drop.
        Similar to the results on MN40, APC achieves the best robustness, as shown in \cref{tab:cls_sonn}.
        In particular, APC yields average adversarial accuracies of 75.8\%, 82.8\%, and 81.9\% on PointNet, PointNet++, and DGCNN, respectively, all representing SOTA performance.
        The remarkable robustness of APC on SONN confirms its practical usability, demonstrating that it remains effective even when evaluated on real-world data.

        Although APC is an input-level defense, which typically offers transferability at the cost of robustness, it still outperforms model-level defenses.
        The transferability of APC is further evaluated in the following section.

    \subsection{Cross-model Transferability}
        To validate transferability of APC, we conduct cross-model generalization experiment.
        As an input-level method, APC performs purification directly on input point clouds and can, in principle, be applied to any model as a plug-and-play module once trained.
        \Cref{tab:cross_mn,tab:cross_sonn} present the results of cross-model tranfer on MN40 and SONN. We report the average adversarial accuracies across all attacks.
        APC is trained with source model using adversarial examples targeting that source model. Then, it is transferred to target models and evaluated on adversarial examples targeting corresponding target model, \ie, both models and adversarial examples are unseen during APC training.
        We compare APC with input-level defenses which have transferability, except CausalPC.
        While CausalPC also manipulates the input point clouds, it requires dataset-specific fine-tuning to adapt to its modified architecture.
        Here, the baselines defenses are model-agnostic, meaning that their performances on target models are independent of the source model.

        On both datasets, APC effectively purifies adversarial examples from various attacks and consistently yields high robustness across diverse source-target pairs, demonstrating strong cross-model generalization. 
        Compared to the model-level defenses in \cref{tab:cls_mn,tab:cls_sonn}, which lack transferability, APC achieves comparable or even superior performance.
        Based on standard recognition and cross-model transfer results, we highlight that APC exhibits notable transferability while maintaining state-of-the-art defense performance.

        \paragraph{Efficiency.}
            In addition to strong transferability across models, APC is also highly efficient.
            We measure the inference time required to purify a single adversarial example as well as the number of parameters of the model used.
            For comparison, we consider network-based methods like APC, \ie, DUP-Net and IF-Defense.
            As shown in \cref{tab:efficiency}, APC shows the lowest inference time (0.002s) and requires the fewest parameters (0.2M), corresponding to only about 0.06\% and 10\% of those of IF-Defense.

        \begin{table}[!t]
            \begin{minipage}[c]{0.4\columnwidth}
                \centering
                \vspace{-3.5mm}
                \caption{Inference time and the number of parameters for each method.}
                \label{tab:efficiency}
                \vspace{-3mm}
                \scriptsize
                \setlength\tabcolsep{0.6em}
                \begin{tabular}{l|ccc}
                    \toprule
                    Metric       & DUP & IF & APC \\
                    \midrule
                    Time (s)     & 0.3 & 3.2 & 0.002 \\
                    \#Params (M) & 0.8 & 2.0 & 0.2 \\
                    \bottomrule
                \end{tabular}
            \end{minipage}
            \hspace{0.02\columnwidth}
            \begin{minipage}[c]{0.6\columnwidth}
                \centering
                \caption{Performance by the number \\ of attacks used in hybrid training}
                \label{tab:ab_hybrid}
                \vspace{-3mm}
                \scriptsize
                \setlength\tabcolsep{0.6em}
                \begin{tabular}{c|cc|cc}
                    \toprule
                    \multirow{2}{*}{\#Attacks} & \multicolumn{2}{c|}{MN40} & \multicolumn{2}{c}{SONN} \\
                    \cmidrule(lr){2-3} \cmidrule(lr){4-5}
                     & Cln. & Adv. & Cln. & Adv. \\
                    \midrule
                    1 & 86.6 & 75.5 & 80.0 & 73.5 \\
                    2 & \textbf{87.1} & 82.2 & 79.8 & 74.5 \\
                    3 & 86.6 & \textbf{84.7} & \textbf{80.6} & \textbf{76.1} \\
                    \bottomrule
                \end{tabular}
            \end{minipage}
        \vspace{-5mm}
        \end{table}

    \subsection{Ablation Studies}
        \paragraph{Effect of Attack Diversity in Hybrid Training.}
            \Cref{tab:ab_hybrid} presents performance under varying numbers of attack types used for hybrid training with PointNet.
            To isolate the effect of the number of attacks, we keep total amount of training data fixed.
            The first and second rows show averages over all combinations, \ie, three results.
            The adversarial accuracy gradually improves as the number of attacks increases, reaching 84.7\% and 76.1\% on MN40 and SONN, respectively.
            Importantly, clean accuracy is preserved, indicating that APC enhances robustness without compromising normal performance.
            Note that even with only two attack types, APC already achieves state-of-the-art performance on both datasets.

        \paragraph{Loss Analysis.}
            \Cref{tab:ab_loss} shows ablation studies on the effectiveness of the geometric consistency loss $\mathcal{L}_{geo}$ and the semantic consistency loss $\mathcal{L}_{sem}$.
            On both datasets, using either loss individually improves performance compared to baseline. 
            The semantic consistency loss contributes more to performance improvement than geometric consistency loss, and combining both yields the best overall performance.

            We compare two distance metrics used in the geometric consistency loss, as shown in \cref{tab:ab_dist}.
            The Hausdorff distance is a similar metric to the Chamfer distance, but instead of averaging the distances of all nearest point pairs, it takes their maximum.
            Both metrics perform almost equally well, and we adopt the Chamfer distance as the default.

        \begin{table}[!t]
        \centering
        \caption{Ablations on loss terms and usage of clean data.}
        \label{tab:ab_loss}
        \begin{tabular}{ccc|cc|cc}
            \toprule
            \multirow{2}{*}{$\mathcal{L}_{geo}$} & \multirow{2}{*}{$\mathcal{L}_{sem}$} & \multirow{2}{*}{Clean} & \multicolumn{2}{c|}{ModelNet40} & \multicolumn{2}{c}{ScanObjectNN} \\
            \cmidrule(lr){4-5} \cmidrule(lr){6-7}
             &  &  & Cln. & Adv. & Cln. & Adv. \\
            \midrule
                   &        & \cmark & 85.4 & 83.6 & 79.3 & 74.5 \\
            \cmark &        & \cmark & 85.7 & 84.0 & 79.3 & 75.0 \\
                   & \cmark & \cmark & 86.5 & 84.6 & 80.0 & 75.5 \\
            \cmark & \cmark &        & 85.6 & 84.5 & 79.0 & 75.6 \\
            \cmark & \cmark & \cmark & \textbf{86.6} & \textbf{84.7} & \textbf{80.6} & \textbf{76.1} \\
            \bottomrule
        \end{tabular}%
        \end{table}

        \begin{table}[!t]
        \centering
        \caption{Comparison of two distance metrics.}
        \label{tab:ab_dist}
        \resizebox{0.475\textwidth}{!}{%
        \begin{tabular}{l@{\hskip 5pt}|c@{\hskip 5pt}c@{\hskip 5pt}c@{\hskip 5pt}|c@{\hskip 5pt}c@{\hskip 5pt}c}
            \toprule
            \multirow{2}{*}{Distance} & \multicolumn{3}{c|}{ModelNet40} & \multicolumn{3}{c}{ScanObjectNN} \\
            \cmidrule(lr){2-4} \cmidrule(lr){5-7}
             & PointNet & PointNet++ & DGCNN & PointNet & PointNet++ & DGCNN \\
            \midrule
            Hausdorff Dist. & 84.5 & 85.1 & 83.4 & 75.3 & 82.8 & 81.8 \\
            Chamfer Dist.   & 84.7 & 84.9 & 83.4 & 76.1 & 82.9 & 81.9 \\
            \bottomrule
        \end{tabular}%
        }
        \end{table}

    \subsection{Clean--Adversarial Accuracy Trade-off}
        Balancing clean accuracy and adversarial robustness is crucial for defense, as the two are generally in a trade-off relationship.
        \Cref{tab:cln_acc} reports the clean accuracy of defense methods across datasets and models.
        Our APC maintains competitive clean accuracy, while achieving state-of-the-art adversarial robustness, and even surpasses vanilla performance on the ScanObjectNN.

        In addition, the ``Clean" column in \cref{tab:ab_loss} compares performance with and without clean examples during training.
        Incorporating clean examples improves not only clean accuracy but also adversarial robustness. 
        We assume that, as clean examples serve as supervision in the training, their inclusion provides information about data distribution that purified examples should have.

        \begin{table}[!t]
        \centering
        \caption{Classification results (\%) of clean examples.}
        \label{tab:cln_acc}
        \resizebox{0.475\textwidth}{!}{%
        \begin{tabular}{l@{\hskip 5pt}|c@{\hskip 5pt}c@{\hskip 5pt}c@{\hskip 5pt}|c@{\hskip 5pt}c@{\hskip 5pt}c}
            \toprule
            \multirow{2}{*}{Defense} & \multicolumn{3}{c|}{ModelNet40} & \multicolumn{3}{c}{ScanObjectNN} \\
            \cmidrule(lr){2-4} \cmidrule(lr){5-7}
             & PointNet & PointNet++ & DGCNN & PointNet & PointNet++ & DGCNN \\
            \midrule
            Vanilla    & 87.6 & 89.1 & 90.1 & 80.4 & 86.5 & 87.4 \\
            \midrule
            SRS        & 86.1 & 87.4 & 79.3 & 78.3 & 85.5 & 76.4 \\
            SOR        & 86.3 & 88.1 & 87.5 & 79.7 & 85.3 & 86.7 \\
            DUP-Net    & 85.5 & 88.0 & 54.2 & 79.1 & 86.2 & 77.2 \\
            IF-Defense & 84.8 & 86.8 & 86.9 & 43.3 & 46.6 & 53.2 \\
            CausalPC   & 84.6 &  -   & 84.2 & 75.7 &  -   & 82.8 \\
            AT         & 85.9 & 88.5 & \textbf{88.8} & 79.3 & 85.5 & 87.3 \\
            HT         & \textbf{86.6} & \textbf{88.9} & 88.2 & 79.1 & 85.8 & 85.7 \\
            APC        & \textbf{86.6} & 87.8 & 87.4 & \textbf{80.6} & \textbf{86.4} & \textbf{87.8} \\
            \bottomrule
        \end{tabular}%
        }
        \end{table}

\begin{figure}
            \centering
                \includegraphics[width=1.\linewidth]{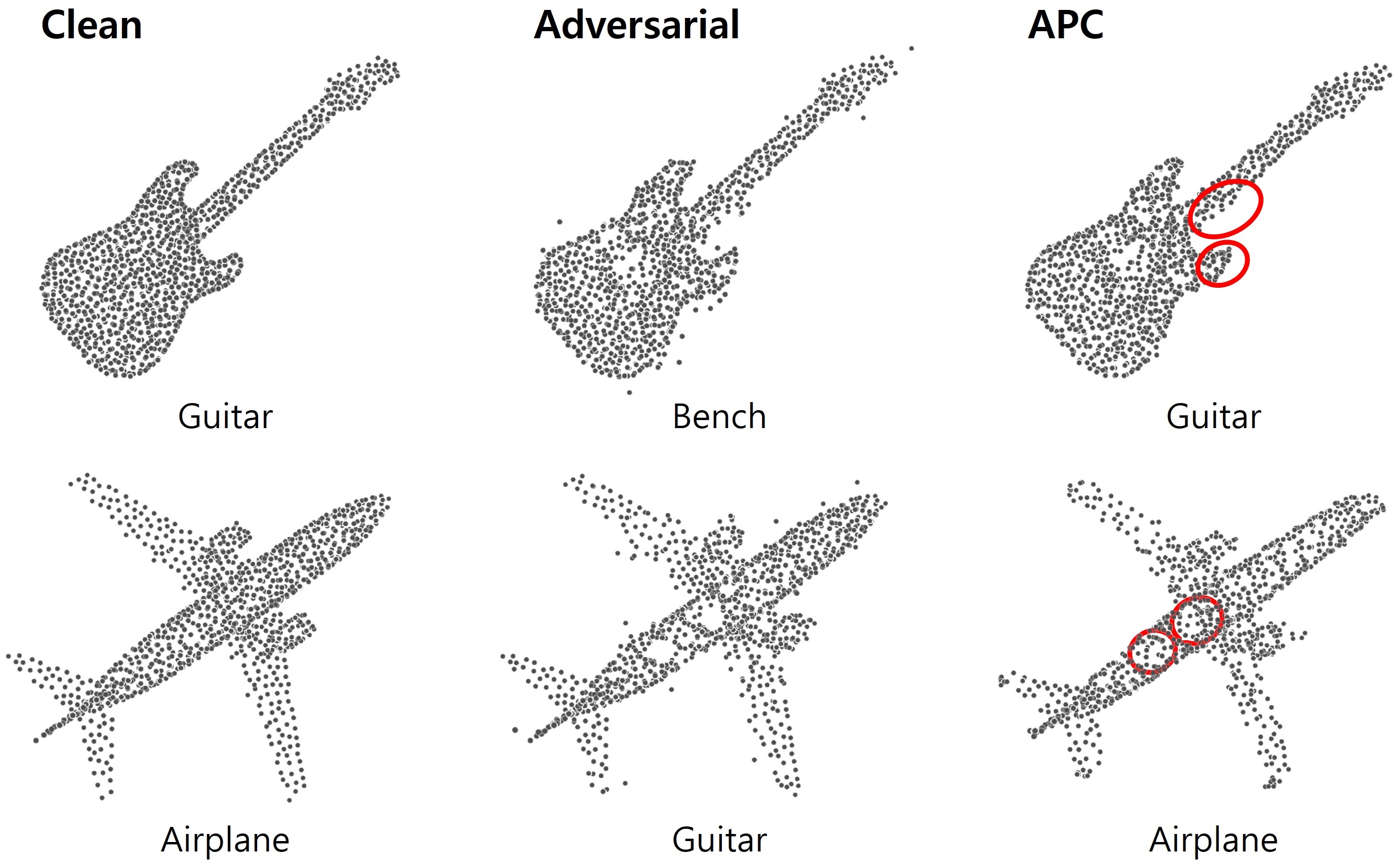}
            \caption{Visualization of clean, adversarial, and purified examples. The labels below each point cloud are the prediction outputs of the victim model.}
            \label{fig:qual}
            \vspace{-1mm}
            \end{figure}

        \paragraph{Visualization.}
            \Cref{fig:qual} visualizes purification results of APC.
            First column shows clean examples correctly classified as ``Guitar" and ``Airplane".
            Adversarial examples are in the middle column, where same objects are misclassified as ``Bench" and ``Guitar", respectively. 
            Third column presents the purified examples after applying APC, successfully recovering the correct labels. 
            Red circles highlight the regions where perturbations are mitigated by restoring displaced points onto object surface and filling empty areas.
            
\section{Conclusion}
\label{sec:con}
     In this paper, we propose Adversarial Point Counterattack (APC), a lightweight input-level purification module that generates per-point counter-perturbations to neutralize adversarial attacks. 
     APC is trained on clean–adversarial pairs with geometric and semantic consistency losses, enabling it to simultaneously recover local geometry and global semantics.
     Through hybrid training over multiple attack types and its input-level design, APC generalizes robustly to unseen attacks and transfers effectively to unseen models.
     Extensive experiments on synthetic and real-scanned 3D recognition benchmarks demonstrate that APC achiev staeste-of-the-art adversarial accuracies.
     Moreover, APC provides this robustness with negligible overhead---requiring only a 0.002s forward pass and 0.2M parameters---making it practical for various real-time 3D applications.

\paragraph{Acknowledgments.} This research was supported by the National Research Foundation of Korea(NRF) grant funded by the korea government(MSIT) for Geunyoung Jung (RS-2022-NR068754) and for Jiyoung Jung (RS-2025-24523036).

{
    \small
    \bibliographystyle{ieeenat_fullname}
    \bibliography{main}
}


\end{document}